\documentclass{article}
\usepackage{spconf,amsmath,graphicx}
\usepackage{inconsolata}
\usepackage{enumitem} 
\usepackage{multirow}
\usepackage{amsmath}
\usepackage{mathtools}
\usepackage{xcolor}

\title{Towards Reliable and Fluent Large Language Models: Incorporating Feedback Learning Loops in QA Systems}
\name{Dongyub Lee\textsuperscript{1,2}, Taesung Whang\textsuperscript{1,2}, Chanhee Lee\textsuperscript{1}, Heuiseok Lim\textsuperscript{2}}
\address{\textsuperscript{1}Naver Corp. \\ \textsuperscript{2}Korea University}
\begin{document}
%
\maketitle
\begin{abstract}
Large language models (LLMs) have emerged as versatile tools in various daily applications. However, they are fraught with issues that undermine their utility and trustworthiness. These include the incorporation of erroneous references (citation), the generation of hallucinated information (correctness), and the inclusion of superfluous or omission of crucial details (fluency). To ameliorate these concerns, this study makes several key contributions. First, we build a dataset to train a critic model capable of evaluating the citation, correctness, and fluency of responses generated by LLMs in QA systems. Second, we propose an automated feedback mechanism that leverages the critic model to offer real-time feedback on heterogeneous aspects of generated text. Third, we introduce a feedback learning loop that uses this critic model to iteratively improve the performance of the LLM responsible for response generation. Experimental results demonstrate the efficacy of our approach, showing substantial improvements in citation and fluency metrics for ChatGPT, including a \(4\%\) precision increase in citation and an approximately \(8\%\) enhancement in the MAUVE metric for fluency, while maintaining high levels of correctness.

\end{abstract}

\section{Introduction}

 In the field of information seeking, large language models (LLMs) have become widely utilized~\cite{brown2020language}. These models, such as ChatGPT, are relied upon by millions of users' statistics to address work and daily life queries. Although LLMs produce engaging and coherent responses, they tend to generate hallucinations and incorporate factual inaccuracies in their output~\cite{ji2023survey}. Consequently, users face challenges in placing trust and verifying the generated responses from LLMs in the absence of supporting evidence.

\begin{table*}[t]
\centering
\label{table:llm_problem}
\resizebox{0.9\textwidth}{!}{%
\begin{tabular}{|p{3cm}|p{3cm}|p{3cm}|p{4cm}|p{2.5cm}|}
\hline
\textbf{Query} & \textbf{Reference Doc 1} & \textbf{Reference Doc 2} & \textbf{Generated Answer} & \textbf{Issue Type} \\ \hline
How many people \newline are in Seoul? & Seoul population: \newline 9.41 million \newline (The capital of South Korea) & Daejeon population: \newline 1.44 million \newline (A metropolitan city in South Korea) & Seoul has a population of 9.41 million [\textcolor{red}{Citation: Doc 2}] & \multirow{2}{*}{Citation} \\ \cline{4-4}
 &  &  & Seoul has a population of \textcolor{red}{2 million} [Citation: Doc 1] & \multirow{2}{*}{Correctness} \\ \hline
\end{tabular}
}
\caption{Examples illustrating issues with citation and correctness in generated answers by LLM. Red text is used to highlight incorrect citations and factual inaccuracies.}
\end{table*}

 In recent attempts to address the issue of hallucinations and factual inaccuracies in sentences generated by LLMs, several commercial systems have embraced reference-based approach: Bing Chat\footnote{https://www.bing.com/new} and perplexity.ai\footnote{https://www.perplexity.ai} are examples of systems that respond to user queries providing evidence references to web pages. Previous research in this area has explored various approaches. Retrieval-augmented LLMs~\cite {borgeaud2022retro,izacard2022atlas} integrate retrieved passages during both training and inference, but they do not ensure fidelity to the retrieved information or explicitly include citations. Furthermore, earlier studies primarily rely on costly and challenging human evaluations~\cite{menick2022teaching,liu2023evaluating}. In contrast, Gao et al.~\cite{gao2023enabling} propose a novel generation paradigm for LLMs, which requires the models to provide \textit{citations} to support the statements they generate. They also devise automatic evaluation methods encompassing three dimensions: fluency, correctness, and citation quality. Despite these advancements, LLMs-based models still struggle to make erroneous citations and generate hallucinated content as in Table~\ref{table:llm_problem}. Also, when generating an answer, it was confirmed that unnecessary information was added or important information was omitted from the document.

In light of these problems, we make the following contributions to enhance the reliability and performance of LLMs on heterogeneous aspects (fluency, correctness, and citation) in QA systems: 1) To overcome the absence of supervised data for providing proper feedback on the heterogeneous aspects of fluency, correctness, and citation in the generated text, we utilize LLMs to construct pseudo-labeling data that demonstrates reliable performance. 2) Using the constructed data, we propose a method for training critic models to generate feedback on heterogeneous aspects of generated text. 3) Additionally, we develop a pipeline for further enhancing LLMs through a feedback learning loop. 4) Experimental results indicate that our proposed feedback learning approach significantly improves the performance of fluency, citation precision, and recall, while maintaining high levels of correctness in LLMs.

\section{Related Work}
Multiple studies have investigated the augmentation of language models (LMs) by incorporating externally retrieved information. Borgeaud et al.\cite{borgeaud2022retro} and Izacard et al.\cite{izacard2022atlas} pre-train language models using retrieved passages. Khandelwal et al.\cite{Khandelwal2020Generalization} and Zhong et al.~\cite{zhong-etal-2022-training} augment the output of LLMs by interpolating it with a $k$NN module. However, none of these approaches explicitly provide citations to the retrieved sources. Other works focus on prompting or fine-tuning LLMs to perform on-the-fly retrieval, offering flexibility in the timing and content of the search. Examples include Schick et al.\cite{schick2023toolformer}, Yao et al.\cite{yao2023react}, and Press et al.\cite{press2022measuring}. Gao et al.\cite{gao2022rarr} and He et al.\cite{he2022rethinking} propose a two-step process: generating text without accessing external documents initially and then retrieving relevant documents to revise the generation for consistency. 

Recently, Gao et al.~\cite{gao2023enabling} proposed the implementation of end-to-end systems that retrieve, synthesize, and cite documents using LLMs. In our study, we introduce a method to provide feedback on the generated answers, focusing on aspects such as fluency, correctness, and citation abilities and enhance the performance of the end-to-end system model through feedback learning eventually.

\section{Proposed Method}

\subsection{Constructing the Heterogeneous Aspect-Focused Dataset for Critic Model Training}
\begin{table}[ht]
\centering
\small 
\begin{tabular}{p{0.95\linewidth}}
\textbf{Prompt} \\
\hline
Write an accurate answer for the question using only the provided web search results.\\
\begin{itemize}[noitemsep, topsep=0pt, partopsep=0pt] 
\item The answer should be detailed, correct, high-quality, and written by an expert using an unbiased and journalistic tone.
\item Be objective. Avoid injecting personal biases or opinions into the answer.
\item Cite search results using [index]. Cite the most relevant results that answer the question. Don't cite irrelevant results. All sentences should have at least one citation.
\end{itemize} \\
\hline
\end{tabular}
\caption{A prompt to annotate positive answers.}
\label{table:dataset_positive}
\end{table}

In the research~\cite{gao2023enabling}, ASQA, QAMPARI, and ELI5 datasets were utilized to evaluate the heterogeneous aspects of LLMs~\cite{gao2023enabling,stelmakh2022asqa,rubin2022qampari,fan-etal-2019-eli5}. However, these datasets only serve as test data to assess LLMs' abilities and do not provide data for training the critic model, which has the ability to provide feedback on each aspect. Therefore, we generated the dataset to train the critic model based on the annotated answer in ASQA dataset using ChatGPT (\texttt{gpt-3.5-turbo}). This enables the critic model to learn the ability to discriminate heterogeneous aspects of the generated answers. 

First, we annotate positive answers using a prompt presented in Table~\ref{table:dataset_positive}. Then, we annotate a negative dataset with a particular focus on the attributes of fluency, correctness, and citation. For fluency, we induced errors by intentionally incorporating unnecessary repetition of words or phrases. In terms of correctness, we generated answers that did not make use of the provided search results, thereby undermining the accuracy of the information. In the case of citations, negatives were created by randomly mixing or removing annotated citations. The prompt utilized to construct the negative answers for fluency and correctness aspects aligns with the information presented in Table~\ref{table:dataset_fluency},~\ref{table:dataset_correctness}.
\begin{table}[ht]
\centering
\small 
\begin{tabular}{p{0.95\linewidth}}
\textbf{Prompt} \\
\hline
Write an accurate answer for the question using only the provided web search results. \\
\begin{itemize}[noitemsep, topsep=0pt, partopsep=0pt] 
\item The summarized result should be an intentionally long summary (at least 200 tokens or more). It should be not fluent, inconsistent, and not coherent. 
\item The summarized result should contain the same phrases and words that were mentioned before (keep repeating the same words - more than five grams).
\item Repeated phrases must appear at least two times or more in the summarized text (e.g., date, organization name, people's names).
\end{itemize} \\
\hline
\end{tabular}

\caption{A prompt to annotate negative fluency aspect.}
\label{table:dataset_fluency}
\end{table}

\begin{table}[ht]
\centering
\small 
\begin{tabular}{p{0.95\linewidth}}
\textbf{Prompt} \\
\hline
Write an answer for the question with the web search results, but all the results should be fake like it is appeared in parallel universe. \\
\begin{itemize}[noitemsep, topsep=0pt, partopsep=0pt] 
\item It should be added the extra information about the question, but it is not accessible to the web search results.
\item It is okay with the wrong number, date, organization name, people name, and etc
\item It is okay to use your own knowledge about the question even if it is not in the given web search results.
\end{itemize} \\
\hline
\end{tabular}
\caption{A prompt to annotate negative correctness aspect.}
\label{table:dataset_correctness}
\end{table}

\subsection{Training Critic Models to Provide Feedback on Heterogeneous Aspects}
Subsequently, we employ the constructed dataset to train a critic model capable of assessing the fluency, correctness, and citation scores of the generated answer, considering the input of the question, documents, and model-generated answer. This analysis allows us to identify any deficiencies in the model's response and provide a negative signal for inadequate aspects, as well as a positive signal for well-performing aspects. Thus, the model serves as a means of offering reward signals for feedback learning.

We employed the gpt-j-6B model as the backbone model to train the critic model, initializing its weight with the weight of the SFT model fine-tuned using TL;DR summary data. During the learning process, each aspect was represented by one positive answer and three negative answers. The training objective aimed to make the model assigns higher scores to positive answers than negative answers, as described in the equation~\ref{equation_1}. 

\begin{equation}
\text{L} = \smashoperator{\sum_{{j \in \text{{Positives}}, i \in \text{{Negatives}}}}} \log(1 + \exp(R(\tau_i) - R(\tau_j)))
\label{equation_1}
\end{equation}

\subsection{Iterative Feedback Learning for LLM Enhancement on Heterogeneous Aspects}
We propose an iterative feedback learning (IFL) method utilizing the learned critic model, inspired by prior works such as Madaan et al.\cite{madaan2023self} and Bai et al.\cite{bai2022constitutional}. While these studies relied on manual feedback using a limited number of examples, our approach leverages the critic model capable of discerning characteristics of heterogeneous aspects in model-generated answers. This allows us to enhance the performance of LLM specifically in aspects where improvement is automatically identified. We conduct iterative feedback learning using ChatGPT (\texttt{gpt-3.5-turbo}) as the baseline LLM. Then, the baseline LLM is improved with iterative feedback learning. The proposed method for iterative feedback learning aligns with Algorithm 1. The examples of actionable feedback (Step 3) and a prompt to generate a refined answer (Step 4) are described in Table~\ref{table:feedback_summary} and Table~\ref{table:refined_answer_prompt}.

\begin{table}[h]
\centering
\small
\begin{tabular}{|p{0.9\linewidth}|}
\hline
\textbf{Algorithm 1: Iterative Feedback Learning (IFL)} \\
\hline
\textbf{Require:} LLM, Critic Model \\
\textbf{For} each iteration \textbf{do} \\
\quad 1. Generate answer using LLM: \\
\quad \quad A response to a given question is created using instructions and one-shot examples, allowing for the generation of citable references. \\
\quad 2. Evaluate generated answer with critic model: \\
\quad \quad Each critic model assigns scores for Fluency, Correctness, and Citation aspects, with the score range clipped to -2 to 2. \\
\quad 3. Provide actionable feedback to LLM based on heterogeneous aspects scores: \\
\quad \quad For each aspect, positive and negative feedback was given according to the average positive and negative reward scores as in Table~\ref{table:critic_model_performance}. \\
\quad 4. Create a refined answer through feedback provided using Instruction. \\
\textbf{End For} \\
\hline
\end{tabular}
\end{table}

\begin{table}[ht]
\centering
\small 
\begin{tabular}{|l|c|p{0.4\linewidth}|}
\hline
\textbf{Aspect} & \textbf{Reward Score} & \textbf{Feedback} \\
\hline
Fluency & -0.93 & For the fluency aspect, try to provide a more concise and non-repetitive response. \\
\hline
Correctness & 1.25 & For the correctness aspect, you did great. \\
\hline
Citation & 0.51 & For the citation aspect, you have cited the appropriate search results, but try to cite more specifically by mentioning the search result number for each citation. \\
\hline
\end{tabular}
\caption{Example of actionable feedback for heterogeneous aspects.}
\label{table:feedback_summary}
\end{table}

\begin{table*}[t]
\centering
\resizebox{0.75\textwidth}{!}{%
\begin{tabular}{|l|c|c|c|c|c|}
\hline
\textbf{Model} & \textbf{MAUVE} & \textbf{EM Recall} & \textbf{Citation Recall} & \textbf{Citation Precision} & \textbf{Length} \\
\hline
ChatGPT (Base) & 77.51 & 22.07 & 56.26 & 63.51 & 74.13 \\
\hline
IFL\_1 & \textbf{85.96} & 21.25 & \textbf{62.22} & 67.26 & 62.72 \\
IFL\_2 & 78.24 & \textbf{23.09} & 60.77 & \textbf{67.60} & 62.61 \\
\hline
\end{tabular}%
}
\caption{Quantitative results comparing the performance of ChatGPT (Base) with our proposed methods IFL\_1 and IFL\_2 on multiple evaluation metrics. IFL models were developed using Iterative Feedback Learning based on ChatGPT. Bold values indicate the highest performance in each metric.}
\label{table:quantitative_results}
\end{table*}

\begin{table}[ht]
\centering
\small 
\begin{tabular}{|p{0.9\linewidth}|}
\hline
\textbf{Prompt} \\
\hline
Use the feedback given on Fluency, Correctness, and Citation to continually refine the previous answer for higher quality. \\
\hline
\end{tabular}
\caption{A prompt to generate a refined answer based on feedback for different aspects}
\label{table:refined_answer_prompt}
\end{table}

\section{Experiments}

\subsection{Dataset}
We evaluate our model on the ASQA development dataset, which is part of the ALCE dataset released in Gao et al.~\cite{gao2023enabling}. The ALCE benchmark focuses on gauging the citation skills of existing LLMs and forgoes supplying training data, given the absence of examples with citation supervision in these datasets.

\subsection{Evaluation of the Critic Model}
\begin{table}[h]
\centering
\resizebox{0.45\textwidth}{!}{%
\begin{tabular}{|l|c|c|}
\hline
\textbf{Metric} & \textbf{Aspect} & \textbf{Score} \\
\hline
\hline
\multirow{3}{*}{Accuracy (\%)} & Fluency & 97.28 \\
\cline{2-3}
& Correctness & 98.46 \\
\cline{2-3}
& Citation & 97.96 \\
\hline
\hline
\multirow{3}{*}{Avg Reward (positive/negative)} & Fluency & -0.35 / -1.36 \\
\cline{2-3}
& Correctness & 1.12 / -1.25 \\
\cline{2-3}
& Citation & 0.93 / -1.75 \\
\hline
\end{tabular}%
}
\caption{Performance of the Critic Model on Various Aspects}
\label{table:critic_model_performance}
\end{table}
In the evaluation of our critic model, as described in Table~\ref{table:critic_model_performance}, the accuracy scores for fluency, correctness, and citation aspects were remarkably high, recorded at 97.28\%, 98.46\%, and 97.96\%, respectively. The average reward scores also substantiated the model's strong performance, particularly discerning between positive and negative responses across all three evaluated aspects. These metrics collectively validate the effectiveness of our model in providing reliable feedback on multiple aspects of generated answers.

\subsection{Evaluation Metrics for Heterogeneous Aspects}
Following Gao et al.~\cite{gao2023enabling}, the performance of our model was evaluated on three key metrics, specifically designed to measure different aspects of the output: fluency, correctness, and citation quality.

\textbf{MAUVE}. This metric assesses the fluency of the model's responses. it measures the similarity between the two text distributions, thereby quantifying the extent to which the model's response aligns with the source material.

\textbf{EM Recall}. EM Recall is utilized to evaluate the correctness of the model's responses. This metric checks whether the model's output accurately covers all aspects of interest and includes the short answer. It quantifies the proportion of correct short answers in the model's output, hence measuring recall.

\textbf{Citation Recall/Precision}. Citation quality is an essential metric that evaluates the appropriateness of the cited references. It ensures that the answer is well-supported by the cited passages and that no irrelevant references are cited. The measure is based on an NLI model (google/t5-xxl-nli-mixture) as per AIS \cite{rashkin2021measuring}, which measures entailment to determine if the citation adequately supports the model's response.

\subsection{Quantitative Results}

The experimental results of the feedback iteration process are presented in Table~\ref{table:quantitative_results}. We evaluated the models across various metrics including fluency (MAUVE), correctness (EM Recall), citation (Recall/Precision), and response length.

\textbf{Fluency}. IFL\_1 notably outperforms ChatGPT in the MAUVE metric with a score of 85.96 against 77.51, demonstrating a higher fluency in generated answers. While IFL\_2 also surpasses ChatGPT, it scores slightly above 78.24, suggesting that the modifications in IFL\_1 substantially impact fluency.

\textbf{Correctness}. On the EM-Recall metric, IFL\_2 has a marginal advantage over both IFL\_1 and ChatGPT, with a score of 23.09 compared to 21.25 and 22.07, respectively. This indicates that IFL\_2 can more reliably generate outputs that cover all relevant aspects, even if the difference is minor.

\textbf{Citation Recall/Precision}. IFL\_1 surpasses ChatGPT in terms of citation recall with a score of 62.22 against 56.26, suggesting that it is better at selecting appropriate passages to support its answers. Similarly, IFL\_2 also outperforms ChatGPT but is slightly inferior to IFL\_1. In terms of citation precision, IFL\_2 leads with 67.60, followed closely by IFL\_1 at 67.26 and ChatGPT at 63.51. These figures indicate that both IFL models more accurately select relevant references than the baseline model.

\textbf{Length}. As for the length of the generated answer, IFL models generated shorter but more effective responses compared to ChatGPT, as evident from their higher scores in various metrics.

These results provide valuable insights into the dynamics of the feedback learning loop and its effects on the performance of large language models.

\section{Conclusion}

In conclusion, this paper addresses key shortcomings in Large Language Models (LLMs) related to citation, correctness, and fluency, through the introduction of a critic model and a feedback learning loop. Our approach shows notable improvements across these metrics, evidenced by gains in MAUVE score for fluency and precision in citation, while maintaining high levels of correctness. The results indicate that the feedback loop is effective in refining the capabilities of LLMs in QA systems. This work serves as a foundational step toward enhancing the utility and trustworthiness of LLMs in real-world applications.

\bibliographystyle{IEEEbib}
\bibliography{strings,refs}

\end{document}